# Searching for Uncollected Litter with Computer Vision

Julian Hernandez
Faculty Mentor: Clark Fitzgerald, Ph.D.

This study combines photo metadata and computer vision to quantify where uncollected litter is present. Images from the Trash Annotations in Context (TACO) dataset were used to teach an algorithm to detect 10 categories of garbage. Although it worked well with smartphone photos, it struggled when trying to process images from vehicle mounted cameras. However, increasing the variety of perspectives and backgrounds in the dataset will help it improve in unfamiliar situations. These data are plotted onto a map which, as accuracy improves, could be used for measuring waste management strategies and quantifying trends.

Keywords – computer vision, trash management

## I. INTRODUCTION

The world is increasingly relying on plastics for food packaging, electronics, single use containers, etc. In the 1960s they made up less than 1% of municipal solid waste, by 2005 this number rose to 10% [1]. Plastics are difficult to recycle, and they take a very long time to break down naturally, meaning that the waste we put into landfills today will be there long into the future. Today only about 10% globally is being recycled, and the rest is either stored in landfills or left to pollute the environment [2].

### A. Harm Caused to Animals

A recently popularized issue with plastic waste is the way that it can harm different animals. Over 270 species have been identified so far that are being harmed or killed by these materials. This is especially prevalent in marine species where over 60% of ocean waste is composed of plastic. When plastic is ingested it can block animals' ability to properly process food, hinder growth, or even cause death. For birds, it can build up in their stomach, and is then passed on to hatchlings in replacement of food, effectively starving them. Sea Turtles are especially affected, with over 86% of their species known to be harmed by plastic waste. Specifically, the Leatherback Sea Turtles are now on the critically endangered list because of how many have died in recent years [2]. The trash that humanity pollutes the ocean with is harming wildlife and destroying ecosystems.

### B. Environmental Consequences

Human produced litter is also contributing to environmental pollution. Post-consumer waste alone accounts for about 5% of total greenhouse gas emissions each year, with methane from poorly managed landfills making up a large proportion of this. When it's not disposed of properly, litter can build up in local waterways, causing flooding and contamination [3]. In Agbogloshie, an E-Waste processing site in Ghana, soil concentrations of lead, barium, copper, and chromium have risen far above safe levels [4]. Because it is cheaper to recycle in countries without environmental protections, it is exported there and processed under poor conditions, leading to a high risk of environmental contamination [5].

C. Societal Costs

Solid waste management costs $200 billion a year, not accounting for the cost of litter that is not probably disposed of. Waste management is usually funded at the municipal level, which isn't an issue for cities in high-income nations. However, in developing nations it can make up the single largest budgetary item for municipalities, taking resources from other valuable programs [3].

Beyond cost, it can cause health risks for both community members and waste management workers. Uncollected litter provides food and shelter to disease carriers such as rodents and insects, allowing their populations to grow quicker and spread further. They can even get into curbside trash waiting to be picked up if it isn't stored in a weatherproof container [3]. Even recyclable materials can cause harm to people if proper precautions are not met. High concentrations of lead have been found in the blood and breastmilk of recyclers working in landfills [4]. Trash management is not only a direct financial burden on governments but can also increase health risks for workers and surrounding communities.

Many of these risks are minimized in high-income nations where environmental and health laws provide protection for consumers and workers. However, to achieve this, these adverse conditions are externalized to low-income nations, through trash and recycling exports. In 2016, 70% of plastic waste from countries within the Organization of Economic Cooperation and Development (OECD) was exported to East Asian countries, primarily China and Indonesia [6]. Europe exports 87% of their plastic to China for recycling, and the United States and Japan rely nearly exclusively on China to recycle their plastic [5]. These exports continue to be financially profitable because of the more relaxed environmental laws in developing nations.

D. Uncollected Litter

In countries where waste collection rates are lower, litter can be a major factor contributing to air and water pollution. Only about 20% trash in the ocean is from fishing and shipping, the other 80% originated on land and eventually washed out there through rivers and streams. A lot of it ends up in the Pacific Ocean Gyre, a collection of garbage that is now twice the size of Texas and extends 100 feet below the ocean surface [2]. To combat this, significant investment needs to be made in waste clean-up activities, especially in highly polluted, low-income nations where trash imports have overwhelmed their trash services [3].

E. Neural Networks

One solution to collect litter before it enters the ocean is using computer vision to automate the process. In 2016, a new algorithm called You Only Look Once (YOLO) provided the first highly accurate, real time object recognition algorithm [7]. This enabled research into using city camera data to identify garbage bags and dumpsters for more efficient collection. [8] A year later another version of YOLO was released that expanded on this, recognizing up to 9000 unique objects [9]. During 2020 a computer vision application detected garbage, leaves, and other debris then cleaned it controlling the equipment on a road sweeper [10]. That same year another algorithm, ResNet, was able to identify trash from multiple types of recycling. These projects have shown the potential for computer vision to aid in the trash management process.

F. Goals

Based on these and other advances in computer vision, an automated trash detection system could help answer these two questions: Where is waste located? How much is there? This is done by leveraging computer vision with images from smartphones and dashcams to analyze specific geographic areas. With only a few dedicated vehicles, daily or weekly, snapshots could be taken to show how much litter there is over time. This information will be used to produce a human readable map that could be used by policy makers and non-profit organizations to evaluate and improve litter reduction programs.

## II. RELATED WORKS

A. Datasets

While there has been a lot of work into recognizing trash using computer vision, these algorithms are trained on small, highly specific datasets that limit their application to the real world. Instead of building on what others have already collected new researchers usually build their own litter image datasets. For example, Yang and Thung addressed the need for specifically trash related image datasets for computer vision by creating one with 2,400 images [11]. Each item was trash and separated into 6 classes: paper, glass, plastic, metal, cardboard, and trash [11]. However, these images only show one item of trash each, and are taken indoors in front of a whiteboard. This prevents computer vision from generalizing for the real world, where trash can be mixed into a large pile and certainly wouldn't be contrasted against a perfectly colored background.

Since then, other researchers have created new datasets with outdoor images that more closely resemble reality. The Trash Annotations in Context (TACO) dataset contains publicly uploaded images of litter that have been outlined and labeled [Fig 1]. It was used in combination with data augmentation to achieve an average accuracy score of 86% [12]. Panwar et al. focused on collecting images of trash in the oceans and underwater and put in a dataset called AquaTrash [14]. With a total of only 369 images, AquaTrash was used to achieve a mean Average Precision (mAP) of 81% when detecting ocean waste [14]. While significant research has been done to label images of trash, there is still a need to vastly increase this over time.

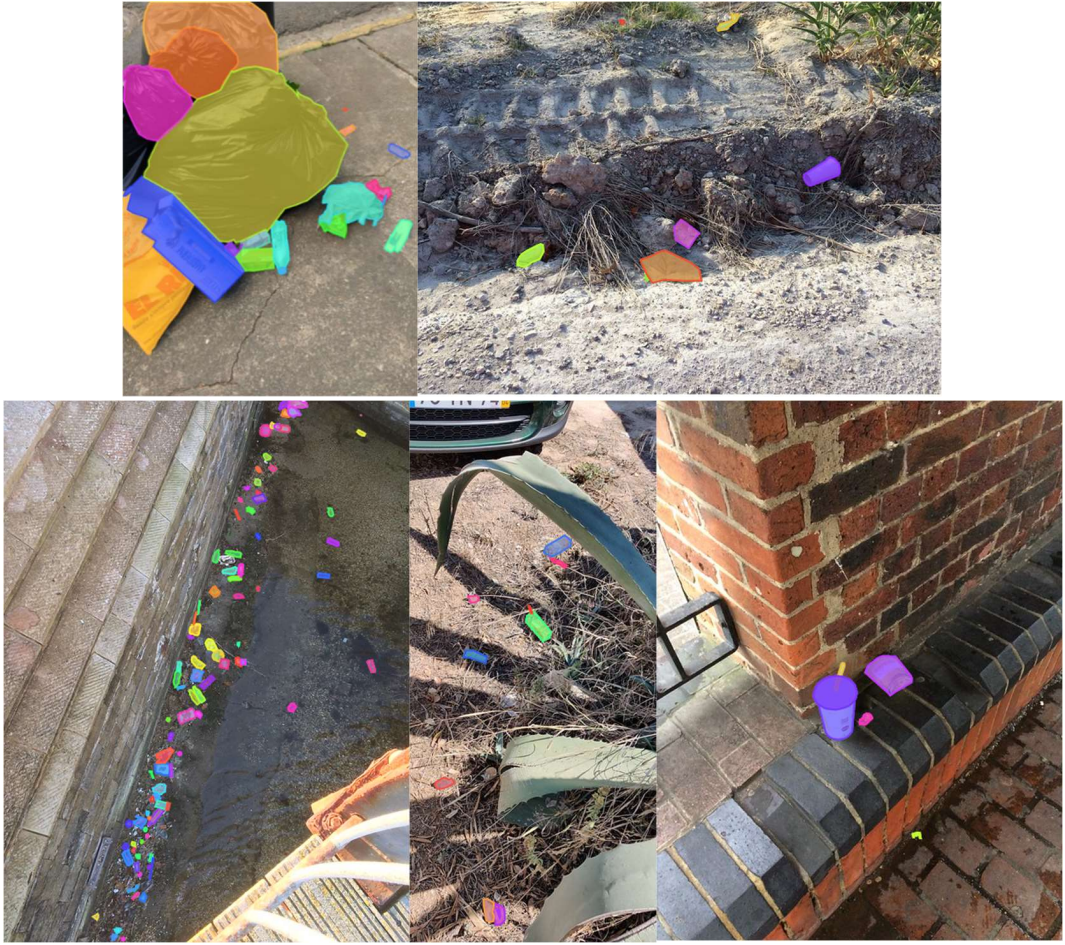

Fig. 1. A sample outlined image from the TACO dataset [15].

B. Classification

An important factor in choosing how to recognize trash is what categories the algorithm will be trained on. Generally, 5 to 20 classes are used, this provides a high level of accuracy while still maintaining differentiation between different categories. Yang and Thung [11] used six classes to sort trash: paper, glass, plastic, metal, cardboard, and trash. According to the EPA, over 50% of current landfill material could be separated into these classes [16]. Others use even more specific categories such as plastic films, cigarettes, garbage bags, and dumpsters [17], [12]. However, a few categories such as wastepaper and plastic bags, consistently had issues with detection because of their non-uniform appearance [17]. Improving image detection categorization also has applications in recycling [18], which benefits from fast, accurate sorting methods [3].

III. METHODOLOGY

Two datasets are needed for detecting objects with computer vision: one for training and one for validation. Traditionally, only one dataset is selected and split into two, with 80% going into a training subset and 20% into a validation subset. Then an algorithm is run which periodically saves a model file that contains the learned information. If training is done correctly then the model will have a generalized concept of each object that it can use to detect them in new images. Because only a few computer vision datasets contained both images of trash and location data, two separate datasets will be used to validate the model.

A. Mask R-CNN

This research uses Masked Regional Convolutional Neural Network (Mask R-CNN), a computer vision algorithm that was made open source by Facebook in 2018 [19]. This extends Faster R-CNN which could only put a box around an object's location, whereas Mask R-CNN can find the exact outline, known as instance segmentation [Fig. 2]. The new algorithm returned more accurate results than previous state of the art instance segmentation algorithms when evaluated on the Common Objects in Context (COCO) dataset, which contains 80,000 images [19].

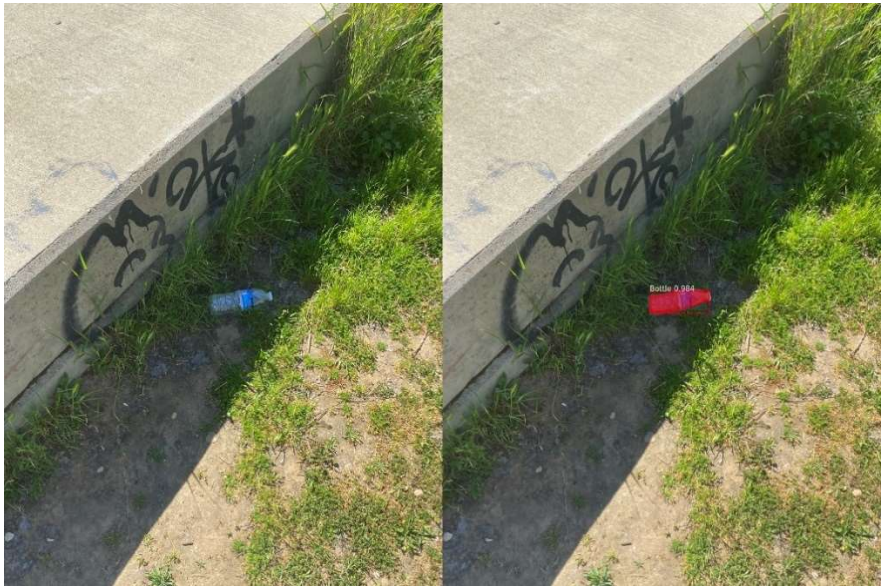

Fig. 2. Mask R-CNN detecting a bottle near California State University, Sacramento.

Mask R-CNN is trained by using specialized computing hardware to repeatedly process large image collections, slowly improving the model's accuracy until it can make new predictions. Features such as shape, color and texture are used to distinguish between different categories of objects. The background features are used to improve the model's understanding of an object's outline as well as rejecting false predictions [19]. Training the model can take multiple weeks to finish even with dedicated servers running around the clock.

The model file remains exceptionally small, around 200MB, even after it has been fully trained. It is then used to make detections, where every pixel in an image is labeled as part of an object or in the background. Although training is a long process, detection only takes a few seconds, and can be run on a small mobile device with fast results [19].

## B. Training Dataset

Because training Mask R-CNN takes such a long time before accurate results are generated a pretrained model provided by the creators of TACO was used. This was trained on their dataset, an open collection of 1500 images of trash with 4,784 individually annotated items. There are 60 different categories, but many of these contain only a few images each so it would be hard for the algorithm to build an abstract understanding for them [Fig 3]. To avoid this the number of unique categories is decreased to 10, the nine largest categories and one new one called *Other Litter*, which combines everything else [15].

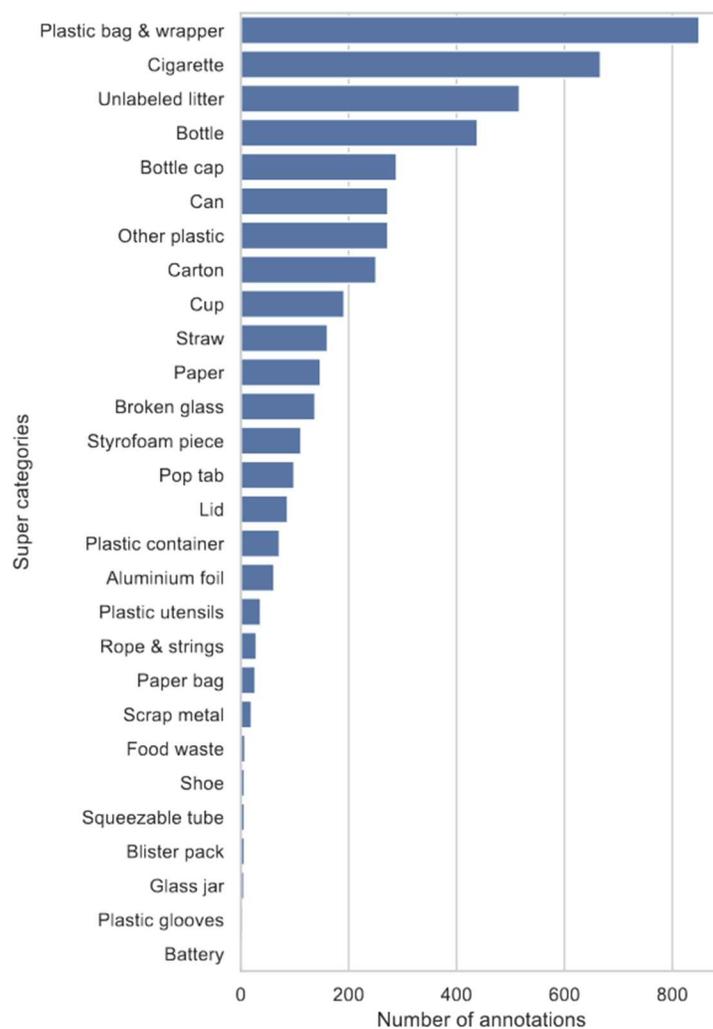

Fig. 3. Number of annotations for each super category in the TACO dataset [15].

## C. Validation Datasets

This model was run on two validation datasets, one with images very similar to the training ones, and another with dashcam images that will test the algorithm from a new perspective.

The first output dataset consisted of hand collected images of trash that were cleaned up at California State University, Sacramento. These were taken using an iPhone 11 before the trash was disposed of. This dataset is representative of what a volunteer organization could collect to estimate categorized totals and gain other insights into their work.

The second validation dataset used vehicle mounted images provided by Mapillary. Mapillary converts community submitted dash cam footage into a collection of images covering the road network, and scrubs any personally identifiable information such as faces and license plates [20]. If detection accuracy is high enough then the data in these images could be quickly processed to locate what streets have the most trash and are most in need cleaning up.

D. Location Estimation

Each detection was placed on a map allowing these data to be accessible to waste management organizations and policy makers in an easily understandable format. The coordinates and camera orientation were extracted from the image's exif metadata and used to estimate the camera viewing angle. Next the outline generated by Mask R-CNN was used to horizontally place that detection into the camera's perspective cone [Fig. 4].

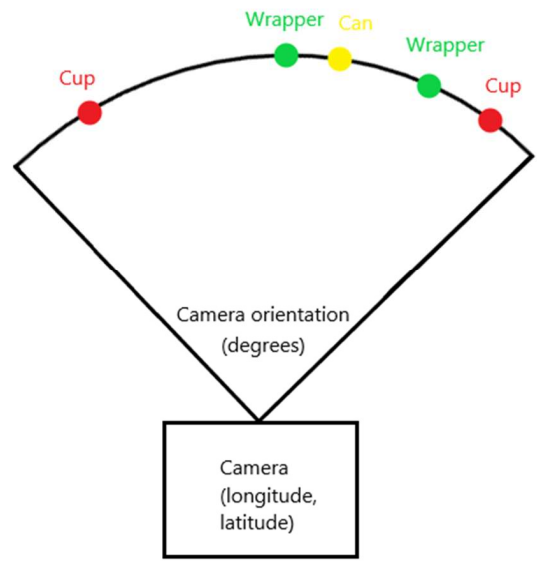

Fig. 4. Camera coordinates and orientation are combined to estimate object location.

IV. RESULTS

A. Metrics

Once trained, evaluation is done to understand how accurately the model can make detections. Mask R-CNN returns a list of trash predictions with how confident it is about each prediction. A confidence threshold is then selected, where predictions above this amount are counted and those below are ignored. A confidence threshold of 30% was used for both output datasets. When these detections are compared against what's actually in the images a confusion matrix can be constructed with four categories: True Positive (TP), False Positive (FP), True Negative (TN), and False Negative (FN) [Table 1].

TABLE 1
CONFUSION MATRIX

|  | Positive | Negative |
|---|---|---|
| True | True Positive (TP) [17] 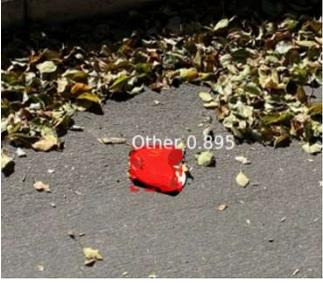 | True Negative (TN) [17] 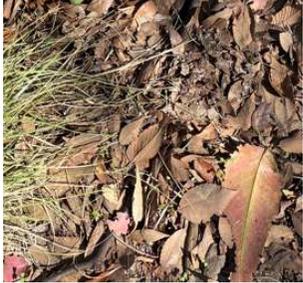 |
| False | False Positive (FP) [17] 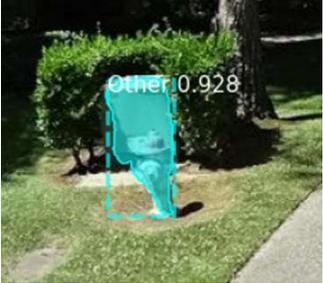 | False Negative (FN) [17] 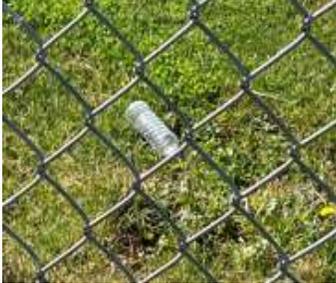 |

Incorrect categorizations were ignored, and any trash detected accurately was counted as a TP. FNs are waste that the model fails to detect. FPs are detections that are labeled as waste but are background objects. TNs are everything else in the image that isn't waste and wasn't predicted to be. TNs aren't quantified because they are not used to calculate accuracy. However, they are important because they are used by the model to reject objects that resemble trash, especially in unfamiliar scenes.

To measure the accuracy of the model, precision and recall are calculated from the total TPs, FNs, and FPs from a sample set of images. Precision conveys what percentage of all predictions were correct. For example, with a precision of 0.9 there's a 90% chance that a randomly chosen prediction is waste and not something else that was misclassified. Recall represents how well the model does at finding all the trash in every image. If detections along a highway had a recall of 0.75, that would mean the model found 75% of the trash along that stretch.

$$Precision = \frac{True\ Positive}{True\ Positive + False\ Positive} \tag{1}$$

$$Recall = \frac{True\ Positive}{True\ Positive + False\ Negative} \tag{2}$$

B. Evaluation

TABLE 2
SMARTPHONE IMAGES CONFUSION MATRIX

|  | Positive | Negative |
|---|---|---|
| True | 145 (TP) | N/A (TN) |
| False | 35 (FP) | 52 (FN) |

Precision: 80%                    Recall: 73%

TABLE 3
MAPILLARY CONFUSION MATRIX

|  | Positive | Negative |
|---|---|---|
| True | 1 (TP) | N/A (TN) |
| False | 98 (FP) | 30 (FN) |

Precision: 1%                    Recall: 3%

When evaluating 150 smartphone images a precision of 80% and recall of 73% [Table 2] are achieved, which is close to results seen by researchers using different algorithms on the same dataset [12]. However, the same model scored 1% and 3% respectively [Table 3] when run on 200 images from Mapillary in the neighborhood surrounding California State University, Sacramento.

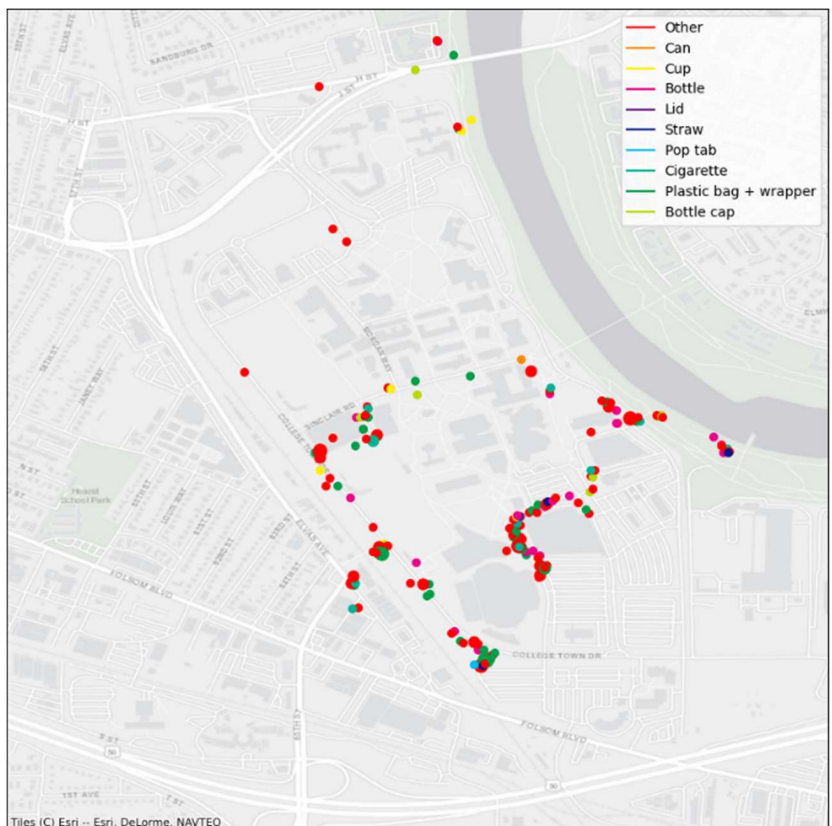
Fig. 5. Detections from smartphone images.

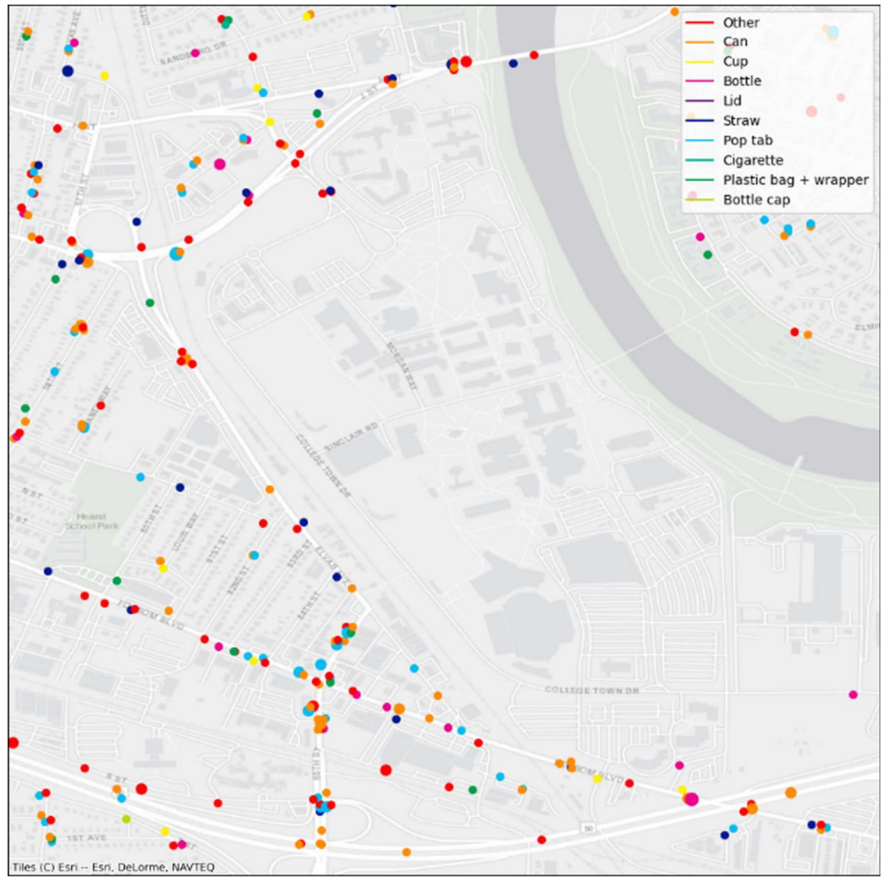

Fig. 6. Detections from Mapillary images.

Combining the detections from the pretrained model and the images' metadata, two maps were created showing the detections from each dataset. Fig. 5 shows the waste cleaned up for the smart phone dataset and where it was found. It clearly outlines the loop that was walked around the main campus buildings, as well as a few scattered detections that come from early testing data. Mapillary image results from the community surrounding the university are shown in Fig. 6. As accuracy improves, computer vision generated maps like this could be used to help organizations plan what areas should be cleaned up.

C. Issues

Although the accuracy on the Mapillary images was low, results from the smartphone images was similar to previous research [12]. This accuracy discrepancy between datasets is likely due to three issues: the camera pixel densities, the training dataset perspective, and misclassified unfamiliar objects.

Modern smartphones have very high-definition cameras which provide a high level of detail and clarity. However, the dashcam images used had a lower pixel count, and because they were originally videos, some details may be lost due to video compression [21]. The model previously struggled with detecting lower resolution objects with less than 20 x 20 pixels. [13]. Cigarettes, for example, have the second most labeled instances, and are generally uniform in color and appearance, so the model should be able to recognize them very easily. However, because of the small size, it had difficulty extracting features it could use to accurately distinguish them from leaves and other background objects [17].

Because the images in the training dataset are mostly smartphone pictures, they are usually taken very close to the objects and from above. This can cause the model to overfit to this perspective, and not generalize its understanding of trash to a horizontal, road facing camera. Additionally, the trash in the Mapillary images tended to be within the range of 10 to 30 feet, making them appear much smaller and decreasing the model's confidence.

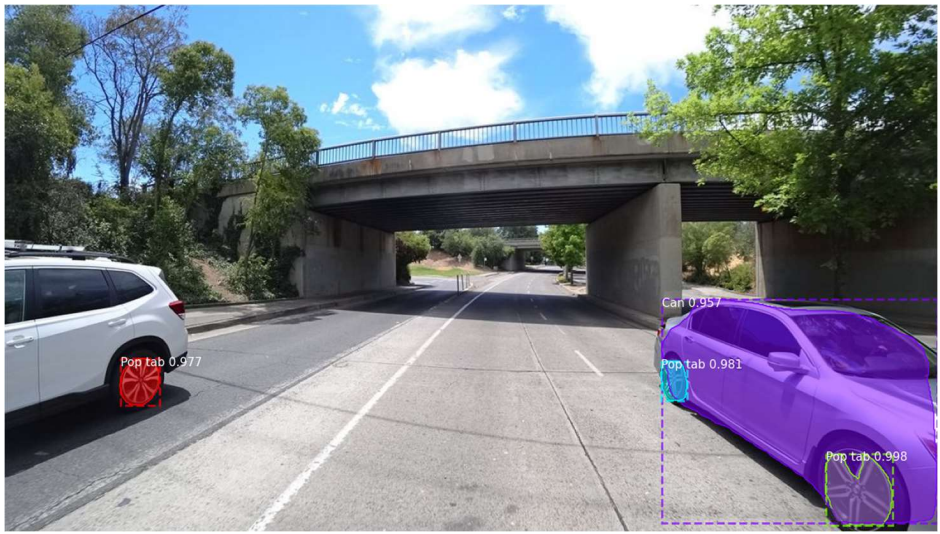

Fig. 7. Cars being recognized as cans with pop tabs for tires by Mask R-CNN. [13]

Lastly, there were specific objects the model hadn't seen during training that it routinely misclassified as trash. The largest cause of FPs was cars being labeled as a can with pop tabs for wheels, which made up 50% of FPs [Fig. 7]. The model could more accurately label cars as TN if the training dataset contained more images with labeled waste in the foreground and cars in the background.

## V. FUTURE WORK AND CONCLUSION

A. Future Work

Several issues were outlined after analyzing results, these will need to be addressed to increase the accuracy to an acceptable level. To fix the pixel density issue a higher resolution dataset could be used, or better cameras could scan and update Mapillary to improve the image detail. Additionally, adding new images to the TACO dataset that are from Mapillary will allow it to learn a more generalized concept of what waste will look like in new perspectives. To improve location estimation multiple images could be combined to triangulate the position [22]. As these improvements are made, more accurate maps of communities could quickly be generated to show where litter is located with a high level of detail.

B. Conclusion

      This research demonstrates how computer vision can be used to aid volunteers in finding what areas have the most litter [Fig. 6] and measuring how much of that waste gets removed during a cleanup [Fig. 5]. These techniques will help reduce the amount of litter in the environment, which has been rapidly increasing due to a combination of high waste production, and low collection rates. This can harm animals [2], be a health risk for surrounding communities [4], and cause environmental hazards [3]. While trash clean-up activities can help collect and dispose of litter, they are expensive, and thus usually only focus on especially dirty areas [23]. Using computer vision to quantify trash levels in different communities would help these organizations focus their efforts where they're most needed. Because global waste production isn't predicted to plateau until the end of the century, new technologies such as this, are needed to help clean up trash faster than it's being littered.